\documentclass[sigconf]{acmart}
\AtBeginDocument{%
  \providecommand\BibTeX{{%
    \normalfont B\kern-0.5em{\scshape i\kern-0.25em b}\kern-0.8em\TeX}}}

\copyrightyear{2022}
\acmYear{2022}
\setcopyright{acmcopyright}\acmConference[KDD '22]{Proceedings of the 28th ACM SIGKDD Conference on Knowledge Discovery and Data Mining}{August 14--18, 2022}{Washington, DC, USA}
\acmBooktitle{Proceedings of the 28th ACM SIGKDD Conference on Knowledge Discovery and Data Mining (KDD '22), August 14--18, 2022, Washington, DC, USA}
\acmPrice{15.00}
\acmDOI{10.1145/3534678.3539366}
\acmISBN{978-1-4503-9385-0/22/08}

\usepackage{colortbl}
\usepackage{algorithmic}
\usepackage[ruled,linesnumbered,vlined]{algorithm2e}

\usepackage{enumitem}
\usepackage{verbatim}
\usepackage{CJKutf8}

\newcommand{\Lapl}{\mathbf{\mathop{\mathcal{L}}}}

\newcommand{\Trans}[1]{{#1}^{\top}}

\newcommand{\Mat}[1]{\mathbf{#1}}

\newcommand{\Space}[1]{\mathbb{#1}}
\newcommand{\Set}[1]{\mathcal{#1}}

\newcommand{\ie}{\textit{i.e., }}
\newcommand{\eg}{\textit{e.g., }}

\newcommand{\wxx}[1]{{\color{black}{#1}}}
\newcommand{\wx}[1]{{\color{black}{#1}}}

\definecolor{syd_color}{rgb}{0.858, 0.188, 0.478}
\newcommand{\syd}[1]{\textcolor{black}{#1}}

\begin{document}
\fancyhead{}
\title{Causal Attention for Interpretable and Generalizable \\ Graph Classification}


\author{Yongduo Sui}
\email{syd2019@mail.ustc.edu.cn}
\affiliation{%
  \institution{University of Science and Technology of China}
  \country{}}

\author{Xiang Wang}
\authornote{Xiang Wang is the corresponding author.}
\email{xiangwang1223@gmail.com}
\affiliation{\institution{University of Science and Technology of China}
  \country{}}

\author{Jiancan Wu}
\email{wujcan@gmail.com}
\affiliation{%
  \institution{University of Science and Technology of China}
  \country{}}

\author{Min Lin}
\email{linmin@sea.com}
\affiliation{%
  \institution{Sea AI Lab}
  \country{}
}

\author{Xiangnan He}
\email{xiangnanhe@gmail.com}
\affiliation{%
 \institution{University of Science and Technology of China}
 \country{}}

\author{Tat-Seng Chua}
\email{dcscts@nus.edu.sg}
\affiliation{%
  \institution{National University of Singapore}
  \country{}}

\renewcommand{\shortauthors}{Sui et al.}

\begin{abstract}
In graph classification, attention- and pooling-based graph neural networks (GNNs) prevail to extract the critical features from the input graph and support the prediction.
They mostly follow the paradigm of ``learning to attend'', which maximizes the mutual information between the attended graph and the ground-truth label.
However, this paradigm makes GNN classifiers recklessly absorb all the statistical correlations between input features and labels in the training data, without distinguishing the causal and noncausal effects of features.
Instead of underscoring the causal features, the attended graphs are prone to visit the noncausal features as the shortcut to predictions.
Such shortcut features might easily change outside the training distribution, thereby making the GNN classifiers suffer from poor generalization.

In this work, we take a causal look at the GNN modeling for graph classification.
With our causal assumption, the shortcut feature serves as a confounder between the causal feature and prediction.
It tricks the classifier to learn spurious correlations that facilitate the prediction in in-distribution (ID) test evaluation, while causing the performance drop in out-of-distribution (OOD) test data.
To endow the classifier with better interpretation and generalization, we propose the \underline{C}ausal \underline{A}ttention \underline{L}earning (CAL) strategy, which discovers the causal patterns and mitigates the confounding effect of shortcuts.
Specifically, we employ attention modules to estimate the causal and shortcut features of the input graph.
We then parameterize the backdoor adjustment of causal theory --- combine each causal feature with various shortcut features.
It encourages the stable relationships between the causal estimation and the prediction, regardless of the changes in shortcut parts and distributions.
Extensive experiments on synthetic and real-world datasets demonstrate the effectiveness of CAL.

\end{abstract}



\maketitle

\section{Introduction}

Graph neural networks (GNNs) \cite{dwivedi2020benchmarking, kipf2016semi} have exhibited impressive performance of graph classification across various domains, such as chemical molecules, social networks, and transaction graphs.
Such a success mainly comes from the powerful representation learning of GNNs, which incorporates the graph structure and encodes them into the representations in an end-to-end way.
Hence, it is crucial to emphasize the critical part of the input graph, while filtering the trivial part out \cite{lin2021generative, DIR, RC-Explainer,ReFine}.
For example, when classifying the mutagenic property of a molecular graph \cite{morris2020tudataset}, GNNs are expected to latch on the functional groups (\ie nitrogen dioxide (NO$_{2}$)), instead of the irrelevant patterns (\ie carbon rings) \cite{debnath1991structure,XGNN};
when detecting fraud in a transaction network, malicious behaviors or coalitions of users are more informative than benign features.


\syd{Towards specifying the critical parts in graphs, some follow-on studies \cite{velivckovic2018graph,kim2020find,gao2019graph,ying2018hierarchical} adopt the paradigm of ``learning to attend'' \cite{DBLP:conf/icml/XuBKCCSZB15,DBLP:conf/nips/VaswaniSPUJGKP17} ---
maximizing the mutual information between the attended graph and the ground-truth label --- to find the attended graph that maximizes the predictive performance.
Specifically, there are two research lines in this paradigm:
(1) Attention-based methods \cite{velivckovic2018graph,kim2020find,brody2021attentive,thekumparampil2018attention,li2015gated}.
They often utilize the attention modules for nodes or edges to locate the attended graphs.
These attention modules act like soft masks to identify the importance of each edge and node to the final representations and predictions.
(2) Pooling-based methods \cite{lee2019self,gao2019graph,ying2018hierarchical,zhang2018end}.
They directly adopt hard masks to select a subset of nodes or edges as the attended graphs, to perform the information propagations.
These attended graphs aim to approach the features that are beneficial for minimizing the training loss, instead of distinguishing the causal and noncausal effects.}



\syd{Unfortunately, recent efforts \cite{geirhos2018imagenet,geirhos2020shortcut,arjovsky2019invariant,knyazev2019understanding} have shown that the current attention or pooling learning methods are prone to exploit the shortcut features to make decisions.}
These shortcuts usually come from the data selection biases, noisy features, or some trivial patterns from graphs, which are noncausal but discriminative in training data.
Due to the existence of these shortcuts, models can capture shortcut features to finish the classification tasks without struggling to learn causal features.
For example, instead of probing into the causal effect of the functional groups, the attended graphs prefer ``carbon rings'' as the cues of the ``mutagenic'' class, because most training ``mutagenic'' molecules are in the ``carbon rings'' context.
While such correlations represent statistical relations inherent in the training data and are beneficial to the in-distribution (ID) test evaluations, they inevitably cause a huge performance drop in the out-of-distribution (OOD) test data \wxx{that are at odds with the training distribution}.
Taking the molecule classification as an example again, when most test ``non-mutagenic'' molecules appear in the ``carbon rings'' context, the attended graphs mislead the GNNs to still predict ``mutagenic''.
As the assumption that the test data conforms to the training distribution is often infeasible in real-world scenarios, the poor generalization of these methods hinders their deployment on critical applications.

To resolve this issue, we first take a causal look at the decision-making process of GNNs for graph classification, which delineates the relationships among the causal feature, shortcut feature, and prediction.
\wxx{With our causal assumption in Figure \ref{fig:causal},} the shortcut feature serves as a confounder \cite{pearl2000models}. 
It opens a backdoor path \cite{pearl2014interpretation} and makes the causal feature and prediction spuriously correlated, \eg misclassifying ``non-mutagenic'' molecules with ``carbon rings'' to the ``mutagenic'' molecules.
Hence, mitigating the confounding effect is promising to exploit the causal features while filtering out the shortcut patterns, thereby enhancing the generalization.

Towards this end, we propose the \underline{C}ausal \underline{A}ttention \underline{L}earning (CAL) strategy  --- 
maximizing the causal effect of the attended graph on predicting the label, while reducing the confounding effect of the shortcut features.
Our attended graph aims to approach the causal features in the graph (\eg nitrogen dioxide), while its complement targets the shortcut features (\eg carbon rings).
Specifically, we first apply attention modules to generate the estimations of the causal and shortcut features from the input graphs.
We then parameterize the backdoor adjustment in the causal theory \cite{pearl2000models,pearl2014interpretation}, which combines each causal estimation with various shortcut estimations and encourages these combinations to maintain a stable prediction.
It encourages the invariant relationships between the causal patterns and the predictions, regardless of the changes in the shortcut parts and distribution shifts.
We apply CAL to various GNN architectures for graph classification.
Experimental results on numerous synthetic and real-world datasets demonstrate the better generalization and insightful interpretations of CAL. 

Our technical contributions are summarized as:
\begin{itemize}[leftmargin=*]
    \item We emphasize the generalization issue of current attention- and pooling-based GNNs in graph classification. 
    From the causal perspective, we ascribe such an issue to the confounding effect of the shortcut features.
    \item \syd{We present a novel Causal Attention Learning (CAL) strategy for graph classification.
    It makes GNNs exploit the causal features while filtering out the shortcut patterns.}
    

    \item Extensive experiments on synthetic and real-world datasets justify the effectiveness of CAL.
    More visualizations with in-depth analyses demonstrate the interpretability and rationality of CAL.
    
\end{itemize}

\section{Preliminaries}
\subsection{Notations}
We denote a graph by $\mathcal{G} = \{\Mat{A}, \Mat{X}\}$ with the node set $\mathcal{V}$ and edge set $\mathcal{E}$.
Let $\Mat{X} \in \mathbb{R}^{|\mathcal{V}| \times \mathrm{F}}$ be the node feature matrix, where $\Mat{x}_i = \Mat{X}[i,:]$ is the $\mathrm{F}$-dimensional attribute vector of node $v_i\in\mathcal{V}$. 
We \wx{use} the adjacency matrix $\Mat{A} \in \mathbb{R}^{|\mathcal{V}| \times |\mathcal{V}| }$ to \wx{delineate} the whole graph structure, where $\Mat{A}[i,j]=1$ if edge $(v_i, v_j) \in \mathcal{E}$\wx{, otherwise} $\Mat{A}[i,j]=0$.
We define $\mathrm{GConv}(\cdot)$ as a GNN layer module and \wxx{denote} the node representation matrix \wxx{by} $\Mat{H}\in\Space{R}^{|\Set{V}|\times d}$, whose $i$-th row $\Mat{h}_{i}=\Mat{H}[i,:]$ denotes the representation of node $v_i$.


\subsection{\syd{Attention Mechanism in GNNs}}
In GNNs, attention can be defined over edges or nodes.
For edge-level attentions \cite{velivckovic2018graph,brody2021attentive,kim2020find,thekumparampil2018attention,lee2019graph}, they utilize weighted message passing and aggregation to update node representations $\Mat{H}'$:
\begin{equation}
\Mat{H}' = \mathrm{GConv}\left(\Mat{A} \odot \Mat{M}_a, \Mat{H}\right)
\label{equ:gnn1}
\end{equation}
where $\Mat{M}_a\in \mathbb{R}^{|\mathcal{V}| \times |\mathcal{V}|}$ denotes the attention matrix that \wxx{is often derived} from trainable parameters and node representations.
For node-level attention, several studies \cite{li2015gated,knyazev2019understanding,lee2019self} define the self-attention mask to select the most attentive node representations:
\begin{equation}
\Mat{H}' = \mathrm{GConv}\left(\Mat{A}, \Mat{H} \odot \Mat{M}_x\right)
\label{equ:gnn2}
\end{equation}
where $\Mat{M}_x\in\Space{R}^{|\Set{V}|\times 1}$ represents the node-level attentions, which can be generated by a network (\eg GNNs or MLPs); $\odot$ is the broadcasted element-wise product. 
Hereafter, we can make further pooling operation \cite{lee2019self} for the output node representations $\Mat{H}^{out}$ and summarize the graph representation $\Mat{h}_\mathcal{G}$ for graph $\mathcal{G}$ via the readout function $f_\mathrm{readout}(\cdot)$.
Then we use a classifier $\Phi$ to project the graph representation into a probability distribution $\Mat{z}_\mathcal{G}$:
\begin{equation}
\Mat{h}_\mathcal{G} = f_\mathrm{readout}\left(\{\Mat{h}_{i}^{out}| i \in \mathcal{V} \}\right), \quad z_\mathcal{G} = \Phi(\Mat{h}_\mathcal{G}).
\end{equation}
These methods follow the paradigm of "learning to attend" by minimizing the following empirical risk:
\begin{equation}
    \mathcal{L}_{\mathrm{CE}} = -\frac{1}{|\mathcal{D}|} \sum_{\mathcal{G}\in\mathcal{D}}\Trans{\Mat{y}}_\mathcal{G}\mathrm{log}(\Mat{z}_\mathcal{G}).
\end{equation}
where $\mathcal{L}_{\mathrm{CE}}$ is the cross-entropy loss over the training data $\mathcal{D}$, and $\Mat{y}_{\mathcal{G}}$ is the ground-truth label vector of $\mathcal{G}$.
However, this learning strategy heavily relies on the statistical correlations between the input graphs and labels. Hence, they will inevitably capture the noncausal shortcut features to make predictions.
\section{Methodology}

In this section, we first analyze the GNN learning from the perspective of causality.
From our causal assumption, we identify the shortcut feature as a confounder.
Then we propose the causal attention learning strategy to alleviate the confounding effect.

\subsection{A Causal View on GNNs}

\begin{figure}[t]
    \centering
    \includegraphics[width=0.9\linewidth]{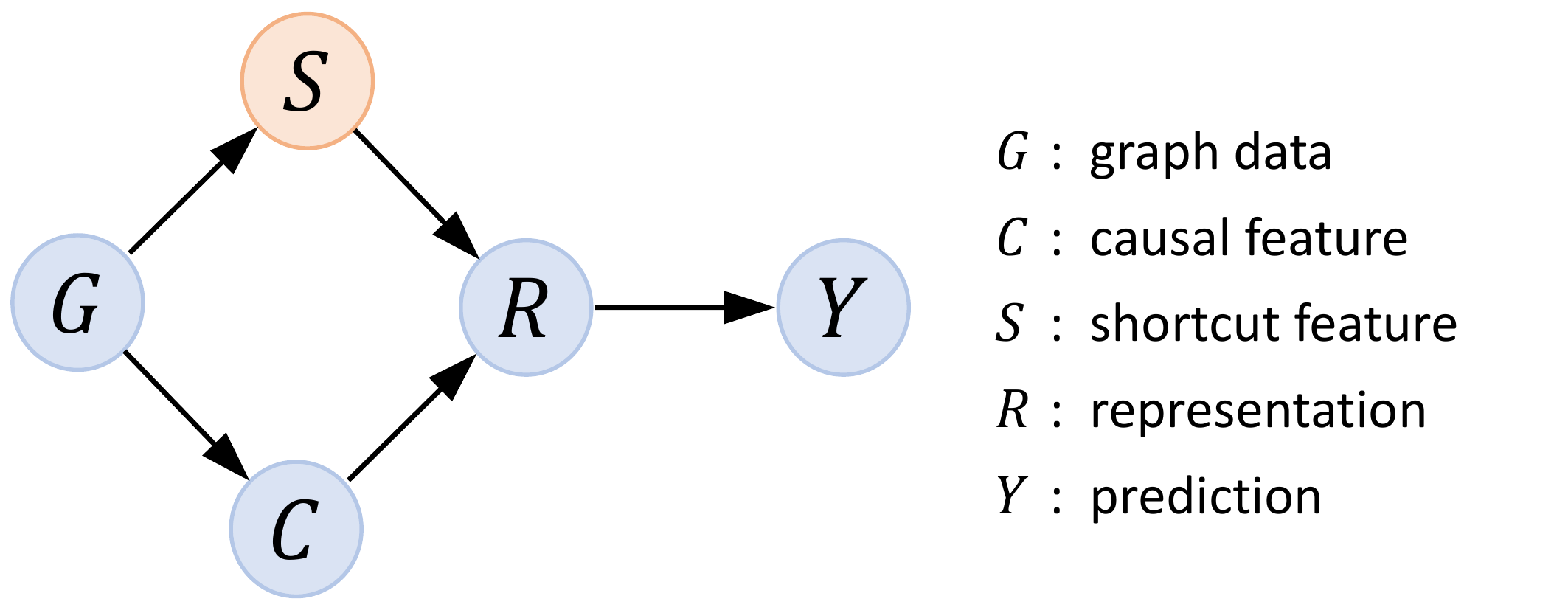}
    \vspace{-2mm}
    \caption{\wxx{Structural causal model for graph classification.}}
    \label{fig:causal}
    \vspace{-4mm}
\end{figure}

We take a causal look at the GNN modeling and construct a \wxx{Structural Causal Model (SCM)} \cite{pearl2000models} in Figure \ref{fig:causal}.
It presents the causalities among five variables: graph data $G$, causal feature $C$, shortcut feature $S$, graph representation $R$, and prediction $Y$, where the link from one variable to another indicates the cause-effect relationship: cause $\rightarrow$ effect. We list the following explanations for SCM:
\begin{itemize}[leftmargin=*]
    \item $\boldsymbol{C\leftarrow G \rightarrow S}$.
    The variable $C$ denotes the causal feature that truly reflects the intrinsic property of the graph data $G$. 
    While $S$ represents the shortcut feature which is usually caused by the data biases or trivial patterns.
    Since $C$ and $S$ naturally coexist in graph data $G$, these causal effects are established.
    
    \item $\boldsymbol{C \rightarrow R \leftarrow S}$.
    The variable $R$ is the representation of the given graph data $G$.
    To generate $R$, the conventional learning strategy \wxx{takes the shortcut feature $S$ and the causal feature $C$ as input to distill discriminative information.}
    
    \item $\boldsymbol{R \rightarrow Y}$.
    The ultimate goal of graph representation learning is to predict the properties of the input graphs.
    The classifier will make prediction $Y$ \wxx{based on} the graph representation $R$.
\end{itemize}    
    
Scrutinizing this SCM, we recognize a backdoor path between $C$ and $Y$, \ie $C \leftarrow G \rightarrow S \rightarrow R \rightarrow Y$, wherein the shortcut feature $S$ \wxx{plays a confounder} role between $C$ and $Y$. Even if $C$ has no direct link to $Y$, the backdoor path will cause $C$ to establish a spurious correlation with $Y$, \eg making wrong predictions based on shortcut feature $S$ instead of causal feature $C$.
Hence, it is crucial to cut off the backdoor path and make the GNN exploit causal features.

\subsection{Backdoor Adjustment}
We have realized that shielding the GNNs from the confounder $S$ is the key to exploiting causal features.
Instead of modeling the confounded $P(Y|C)$ in Figure \ref{fig:causal}, we should achieve the graph representation learning by eliminating the backdoor path.
But how to achieve this?
Fortunately, causal theory \cite{pearl2000models, pearl2014interpretation} provides us with a feasible solution:
we can exploit the $\textbf{do}$-\textbf{calculus} on the variable $C$ to remove the backdoor path by estimating $P_m(Y|C)=P(Y|do(C))$. 
It needs to stratify the confounder $S$ between $C$ and $Y$. 
Therefore, we can obtain the following three essential conclusions:
\begin{itemize}[leftmargin=*]
    \item The marginal probability $P(S = s)$ is invariant under the intervention, because the shortcut feature will not be affected by cutting off the backdoor path. Thus, $P(s) = P_m(s)$.
    \item The conditional probability $P(Y|C,s)$ is invariant, because $Y$’s response to $C$ and $S$ has nothing to do with the causal effect between $C$ and $S$. Then we can get: $P_m(Y|C, s) = P(Y|C, s)$.
    \item Obviously, the variables $C$ and $S$ are independent under the causal intervention, which we have: $P_m(s|C)=P_m(s)$.
\end{itemize}
Based on the above conclusions, we have:
\begin{equation}
\begin{aligned}
\label{equ:backdoor}
    P(Y|do(C)) & = P_m(Y|C) \\
    & = \sum\nolimits_{s\in\mathcal{T}}P_m(Y|C, s)P_m(s|C) \quad (Bayes\,Rule) \\
    & = \sum\nolimits_{s\in\mathcal{T}}P_m(Y|C, s)P_m(s) \quad (Independency)\\
    & = \sum\nolimits_{s\in\mathcal{T}}P(Y|C, s)P(s),
\end{aligned}
\end{equation}
where $\mathcal{T}$ denotes the confounder set; 
$P(Y|C, s)$ represents the conditional probability given the causal feature $C$ and confounder $s$; 
$P(s)$ is the prior probability of the confounder. Equation~\eqref{equ:backdoor} is usually called \textbf{backdoor adjustment} \cite{pearl2014interpretation}, which is a powerful tool to eliminate the confounding effect. 
However, there exist two challenges for implementing Equation~\eqref{equ:backdoor}:
i) The confounder set $\mathcal{T}$ is commonly unobservable and hard to obtain.
ii) Due to the discrete nature of graph data, it seems difficult to \wxx{directly manipulate the graph data, conditioning on domain-specific constraints (\eg valency rules in molecule graphs).}
In section \ref{subs:causal}, we will introduce a simple yet effective solution to overcome these issues.

\subsection{Causal and Trivial Attended-graph}\label{sec31}

Given a graph $\mathcal{G} = \{\Mat{A}, \Mat{X}\}$, we formulate the soft masks on the graph structure and node feature as $\Mat{M}_a\in \mathbb{R}^{|\mathcal{V}| \times |\mathcal{V}|}$ and $\Mat{M}_x\in \mathbb{R}^{|\mathcal{V}| \times 1}$, respectively.
\wx{Wherein, each element of the masks indicates the attention score relevant to the task of interest, which often falls into the range of $(0,1)$.}
Given an arbitrary mask $\Mat{M}$, we define its complementary mask as $\overline{\Mat{M}}=\Mat{1}-\Mat{M}$, where $\Mat{1}$ is \wx{the all-one matrix.}
Therefore, we can divide the full graph $\mathcal{G}$ into two attended-graphs: $\mathcal{G}_1=\{\Mat{A}\odot \Mat{M}_a, \Mat{X}\odot \Mat{M}_x\}$ and $\mathcal{G}_2=\{\Mat{A}\odot \wx{\overline{\Mat{M}}_a, \Mat{X}\odot \overline{\Mat{M}}_x\}}$.

With the inspection on the data-generating process, recent studies \cite{DIR, lin2021generative, ying2019gnnexplainer,knyazev2019understanding} argue that the label of a graph is usually determined by its causal part.
Considering a molecular graph, its mutagenic property relies on the existence of relevant functional groups \cite{RC-Explainer};
Taking the digit image in the form of superpixel graph as another example, the coalition of digit-relevant nodes determines its label.
Formally, given a graph $\Set{G}$, we define the attended graph collecting all causal features as the \textbf{causal attended-graph} $\mathcal{G}_c$, while the counterpart forms the \textbf{trivial attended-graph} $\mathcal{G}_t$.
However, the ground-truth attended-graph is usually unavailable in real-world applications.
Hence, we aim to capture the causal and trivial attended-graph from the full graph by learning the masks: $\mathcal{G}_c=\{\Mat{A}\odot \Mat{M}_a, \Mat{X}\odot \Mat{M}_x\}$ and $\mathcal{G}_t=\{\Mat{A}\odot \overline{\Mat{M}}_a, \Mat{X}\odot \overline{\Mat{M}}_x\}$.
Learning to identify causal attended-graphs not only guides the representation learning of GNNs, but also answers "What knowledge does the GNN use to make predictions?", which is crucial to the applications on explainability, privacy, and fairness.

\begin{figure}[t]
    \centering
    \includegraphics[width=1\linewidth]{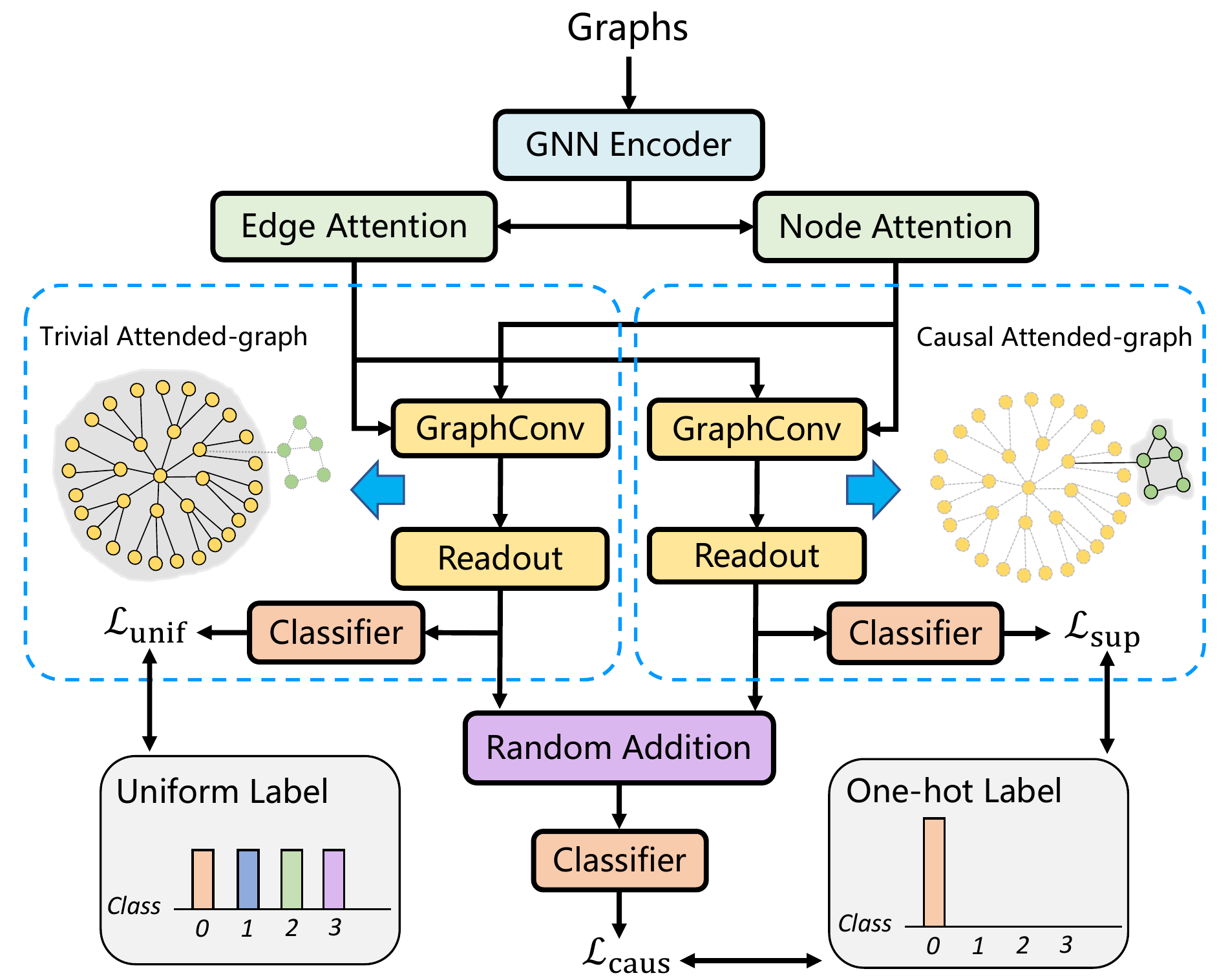}
    \vspace{-4mm}
    \caption{The overview of the proposed Causal Attention Learning (CAL) framework.}
    \label{fig:model}
    \vspace{-4mm}
\end{figure}

\subsection{Causal Attention Learning} \label{sec34}

To implement the aforementioned backdoor adjustment, we propose the \textbf{C}ausal \textbf{A}ttention \textbf{L}earning (CAL) framework:

\subsubsection{\textbf{Estimating soft masks.}}
Towards effective causal intervention, it is necessary to separate the causal and shortcut features from the full graphs.
To this end, we hire attention modules, which yield two branches for the causal and trivial proposals.
Given a GNN-based encoder $f(\cdot)$ and a graph $\mathcal{G}=\{\Mat{A}, \Mat{X}\}$, we can obtain the node representations:
\begin{equation}
\Mat{H}=f(\Mat{A}, \Mat{X}).
\end{equation}
Then we adopt two MLPs: $\mathrm{MLP_{node}}(\cdot)$ and $\mathrm{MLP_{edge}}(\cdot)$ to estimate the attention scores from two orthogonal perspectives: node-level and edge-level. For node $v_i$ and edge $(v_i, v_j)$ we can obtain:
\begin{equation}
\alpha_{c_i}, \alpha_{t_i}=\sigma(\mathrm{MLP_{node}}(\Mat{h}_i)),
\label{equ:node_att}
\end{equation}
\begin{equation}
\beta_{c_{ij}}, \beta_{t_{ij}}=\sigma(\mathrm{MLP_{edge}}(\Mat{h}_i||\Mat{h}_j)),
\label{equ:edge_att}
\end{equation}
where $\sigma(\cdot)$ is softmax function, $||$ denotes concatenation operation;
$\alpha_{c_i}, \beta_{c_{ij}}$ represent the node-level attention score for node $v_i$ and edge-level attention score for edge  $(v_i, v_j)$ in causal attended-graph;
analogously, $\alpha_{t_i}, \beta_{t_{ij}}$ are for trivial attended-graph.
Note that $\alpha_{c_i} + \alpha_{t_i}=1$, and $\beta_{c_{ij}} + \beta_{t_{ij}}=1$.
These attention scores indicate how much the model pays attention to each node or edge in the corresponding attended-graph.
Now we can construct the soft masks $\Mat{M}_x$, $\overline{\Mat{M}}_x$, $\Mat{M}_a$, and $\overline{\Mat{M}}_a$ based on the attention scores $\alpha_{c_i}$, $\alpha_{t_i}$, $\beta_{c_{ij}}$, and $\beta_{t_{ij}}$, respectively.
Finally, we can decompose the original graph $\mathcal{G}$ into the initial causal and trivial attended-graphs: $\mathcal{G}_c=\{\Mat{A}\odot \Mat{M}_a, \Mat{X}\odot \Mat{M}_x\}$ and $\mathcal{G}_t=\{\Mat{A}\odot \overline{\Mat{M}}_a, \Mat{X}\odot \overline{\Mat{M}}_x\}$.

\subsubsection{\textbf{Disentanglement.}}
Until now, we have distributed the attention scores at the granularity of nodes and edges to create the initial attended-graphs. 
Now we need to make the causal and trivial attended-graphs to capture the causal and shortcut features from the input graphs, respectively.
Specifically, we adopt two GNN layers to obtain the representations of attended-graphs and make predictions via readout function and classifiers:
\begin{equation}\label{equ:zc}
\Mat{h}_{\mathcal{G}_c} = f_{\mathrm{readout}}(\mathrm{GConv}_c(\Mat{A}\odot \Mat{M}_a, \Mat{X}\odot \Mat{M}_x)), \quad \Mat{z}_{\mathcal{G}_c} = \Phi_{c}(\Mat{h}_{\mathcal{G}_{c}}),
\end{equation}
\begin{equation}\label{equ:zt}
\Mat{h}_{\mathcal{G}_t} = f_{\mathrm{readout}}(\mathrm{GConv}_t(\Mat{A}\odot \overline{\Mat{M}}_a, \Mat{X}\odot \overline{\Mat{M}}_x)), \quad \Mat{z}_{\mathcal{G}_t} = \Phi_{t}(\Mat{h}_{\mathcal{G}_{t}}).
\end{equation}
The causal attended-graph aims to estimate the causal features, so we classify its representation to the ground-truth label. 
Thus, we define the supervised classification loss as:
\begin{equation}
    \mathcal{L}_{\mathrm{sup}} = -\frac{1}{|\mathcal{D}|} \sum_{\mathcal{G}\in\mathcal{D}}\wx{\Trans{\Mat{y}}_\mathcal{G}}\mathrm{log}(\Mat{z}_{\mathcal{G}_c}).
    \label{equ:loss1}
\end{equation}
where $\mathcal{L}_{\mathrm{sup}}$ is the cross-entropy loss over the training data $\mathcal{D}$. 
The trivial attended-graph aims to approach the trivial patterns that are unnecessary for classification. 
Hence, we push its prediction evenly to all categories and define the uniform classification loss as:

\begin{equation}
    \mathcal{L}_{\mathrm{unif}} = \frac{1}{|\mathcal{D}|} \sum_{\mathcal{G}\in\mathcal{D}}\mathrm{KL}(\Mat{y}_\mathrm{unif}, \Mat{z}_{\mathcal{G}_t}).
    \label{equ:loss2}
\end{equation}
where $\mathrm{KL}$ denotes the KL-Divergence, $\Mat{y}_\mathrm{unif}$ represents the uniform distribution.
By optimizing the above two objectives, we can effectively disentangle causal and trivial features.
Please note that prior efforts \cite{ReFine,RC-Explainer,lin2021generative, ying2019gnnexplainer,knyazev2019understanding} have shown that the mutual information between the causal part and label is greater than that between the full graph and label, due to the widespread trivial patterns or noise.
Hence, the proposed disentanglement will not make the captured causal attended-graph converge to the full graph (noiseless full graph is a special case), which is not an optimal solution.
See Section \ref{sec45} for more supporting evidence and analyses.

\subsubsection{\textbf{Causal intervention}.}\label{subs:causal}
As shown in Equation \eqref{equ:backdoor}, \wxx{one promising solution to alleviating the confounding effect is the backdoor adjustment} --- that is, stratifying the confounder and pairing the target causal attended-graph with every stratification of trivial attended-graph to compose the ``intervened graphs''.
However, due to the irregular graph data, it is impossible to make the intervention on data-level, \eg changing a graph's trivial part to generate a counterfactual graph data.
Towards this end, we make the implicit intervention on representation-level and propose the following loss guided by the backdoor adjustment:
\begin{gather}
    \Mat{z}_{\Set{G}'} = \Phi(\Mat{h}_{\Set{G}_{c}} + \Mat{h}_{\Set{G}_{t'}}),
    \label{equ:inter}
\end{gather}
\begin{gather}
    \Lapl_{\text{caus}} = -\frac{1}{|\Set{D}|\cdot|\hat{\Set{T}}|} \sum_{\Set{G}\in\Set{D}}\sum_{t'\in\hat{\Set{T}}}\Trans{\Mat{y}}_{\Set{G}}\log{(\Mat{z}_{\Set{G}'})},
    \label{equ:loss3}
\end{gather}
where $\Mat{z}_{\Set{G}'}$ is the prediction from a classifier $\Phi$ on ``implicit intervened graph'' $\Set{G}'$;
$\Mat{h}_{\Set{G}_{c}}$ is the representation of causal attended-graph $\Set{G}_{c}$ derived from Equation \eqref{equ:zc};
while $\Mat{h}_{\Set{G}_{t'}}$ is the representation of stratification $\Set{G}_{t'}$ obtained via Equation \eqref{equ:zt};
$\hat{\Set{T}}$ is the estimated stratification set of the trivial attended-graph, which collects the appearing trivial features from training data.
In practice, we apply random addition to make the intervention in Equation \eqref{equ:inter}.
We define the Equation \eqref{equ:loss3} as the causal intervention loss. It pushes the predictions of such intervened graphs to be invariant and stable across different stratifications, due to the shared causal features.
Finally, the objective of CAL can be defined as the sum of the losses:
\begin{equation}
    \mathcal{L}= \mathcal{L}_{\mathrm{sup}} + \lambda_1\mathcal{L}_{\mathrm{unif}} + \lambda_2\mathcal{L}_{\mathrm{caus}}
    \label{equ:loss_all}
\end{equation}
where $\lambda_1$ and $\lambda_2$ are hyper-parameters that determine the strength of disentanglement and causal intervention, respectively.
The detailed algorithm of CAL is provided in Appendix \ref{app1}, Alg.\ref{alg1}, and the overview of CAL is depicted in Figure \ref{fig:model}.

\section{Experiments}
To verify the superiority and effectiveness of the proposed CAL, we conduct experiments to answer the following research questions:

\begin{itemize}[leftmargin=*]
\item \textbf{RQ1:} How effective is the proposed CAL in alleviating the out-of-distribution (OOD) issue?
\item \textbf{RQ2:} Can the proposed CAL achieve performance improvements on real-world datasets?
\item \textbf{RQ3:} For the different components in CAL, what are their roles and impacts on performance?
\item \textbf{RQ4:} Does CAL capture the causal attended-graphs with significant patterns and insightful interpretations?
\end{itemize}

\subsection{Experimental Settings}
\subsubsection{\textbf{Datasets}.}
\begin{figure}[t]
    \centering
    \includegraphics[width=0.95\linewidth]{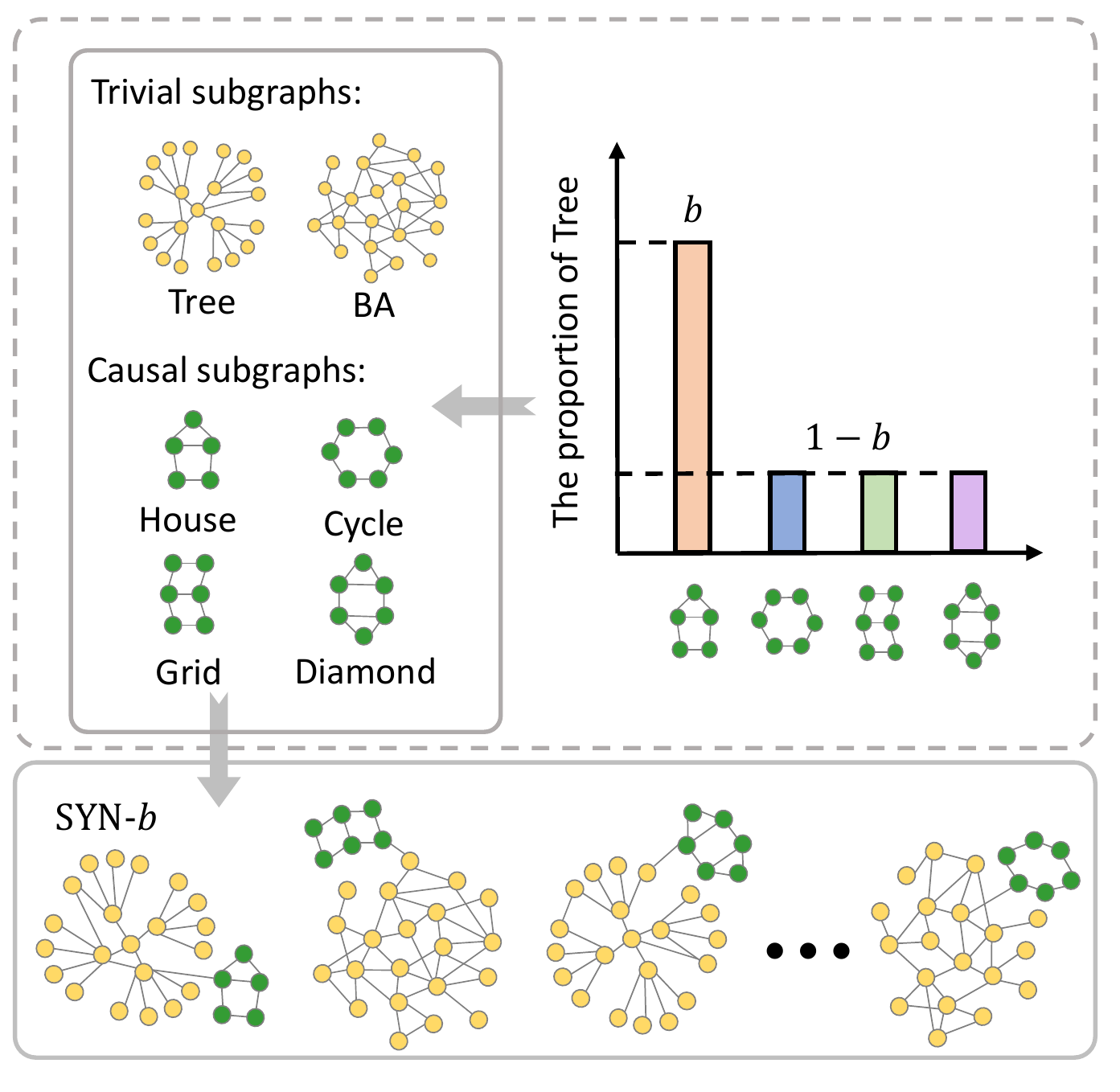}
    \vspace{-4mm}
    \caption{Illustration of the synthetic datasets.}
    \vspace{-4mm}
    \label{fig:syn_dataset}
\end{figure}

We conduct experiments on both synthetic datasets and real-world datasets.
\begin{itemize}[leftmargin=*]
    \item \textbf{Synthetic graphs:} 
    Following \cite{ying2019gnnexplainer}, we create the synthetic dataset for graph classification, which contains a total of 8,000 samples with 4 classes, and keeps balance (2,000 samples) for each class.
    As shown in Figure \ref{fig:syn_dataset}, each sample consists of two parts: causal subgraph and trivial subgraph. More details about the causal and trivial subgraph are provided in Appendix \ref{app2}.
    The task is to predict the type of the causal part in the whole graph.
    For simplicity, we choose the ``House'' class to define the bias-level:
    \begin{equation}
        b = \frac{\# \mathrm{\text{Tree-House}}}{\# \mathrm{House}}
        \label{equ:bias}
    \end{equation}
    where $\#\mathrm{\text{Tree-House}}$ denotes the number of ``House'' causal subgraphs with the ``Tree'' trivial subgraphs, and $\# \mathrm{House}$ presents the number of graphs in the ``House'' class, which is 2,000.
    We set the proportion of ``Tree'' in the other three classes to $1 - b$.
    Obviously, for the unbiased dataset, $b=0.5$.
    We abbreviate the synthetic dataset with bias-level $b$ as SYN-$b$.
    We keep the same bias-level on the training/validation set and keep the testing set unbiased. 
    Please refer to Appendix \ref{app2} for more details.

    \item \textbf{Real-world graphs:} 
    We conduct experiments on three biological datasets (MUTAG, NCI1, PROTEINS), three social datasets (COLLAB, IMDB-B, IMDB-M) \cite{morris2020tudataset}, and two superpixel datasets (MNIST, CIFAR-10) \cite{knyazev2019understanding}. More details, such as statistics and splitting of datasets, are provided in Appendix \ref{app2}.
\end{itemize}

\subsubsection{\textbf{Baselines.}}
To verify the superiority of CAL, we adopt the following prevalent graph classification solutions as baselines:
\begin{itemize}[leftmargin=*]
    \item \textbf{Attention-based methods:} GAT \cite{velivckovic2018graph}, GATv2 \cite{brody2021attentive}, SuperGAT \cite{kim2020find}, GlobalAttention \cite{li2015gated}, AGNN \cite{thekumparampil2018attention}. 
    \item \textbf{Pooling-based methods:} SortPool \cite{zhang2018end}, DiffPool \cite{ying2018hierarchical}, Top-$k$ Pool \cite{gao2019graph}, SAGPool \cite{lee2019self}.
    \item \textbf{Kernel-based methods:} Graphlet kernel (GK) \cite{shervashidze2009efficient}, Weisfeiler Lehman Kernel (WL) \cite{shervashidze2011weisfeiler}, Deep Graph kernels (DGK) \cite{yanardag2015deep}.
    \item \textbf{GNN-based methods:} GCN \cite{kipf2016semi}, GIN \cite{xu2018how}
\end{itemize}
\syd{Besides these methods, we also consider the state-of-the-art algorithms: IRM \cite{arjovsky2019invariant} and DRO \cite{sagawa2019distributionally}, which are particularly designed for OOD issues. Please note that these methods require specific environments or group annotations for each training example, therefore we consider them as the methods with upper bound performance.}

\subsubsection{\textbf{Hyper-parameters}.} All training hyper-parameters and model configurations are summarized in Appendix \ref{app3}. Codes are released at \url{https://github.com/yongduosui/CAL}.

\begin{table*}[htb]
\centering
\caption{Test Accuracy (\%) of graph classification on synthetic datasets with diverse biases. The number in brackets represents the performance degradation compared with the unbiased dataset. Our methods are highlighted with a gray background.}
\label{table:main_syn}
\setlength{\tabcolsep}{5.4mm}{\begin{tabular}{l|ccccc}
\toprule
Method & SYN-$0.1$  & SYN-$0.3$  & Unbiased & SYN-$0.7$ & SYN-$0.9$ \\ 
\toprule
GATv2 \cite{brody2021attentive}  & 87.25 ($\downarrow 7.37\%$) & 92.19 ($\downarrow 2.12\%$) & 94.19 & 93.31 ($\downarrow 0.93\%$) & 90.62 ($\downarrow 3.79\%$) \\
SuperGAT \cite{kim2020find}  & 83.81 ($\downarrow 12.75\%$) & 91.94 ($\downarrow 4.29\%$) & 96.06 & 88.50 ($\downarrow 7.89\%$) & 82.81 ($\downarrow 13.79\%$) \\
GlobalAtt \cite{li2015gated} & 87.19 ($\downarrow 10.40\%$) & 93.75 ($\downarrow 3.66\%$) & 97.31 & 94.62 ($\downarrow 2.76\%$) & 91.50 ($\downarrow 5.97\%$) \\
AGNN \cite{thekumparampil2018attention}  & 84.56 ($\downarrow 11.69\%$) & 93.06 ($\downarrow 2.81\%$) & 95.75 & 94.81 ($\downarrow 0.98\%$) & 88.12 ($\downarrow 7.97\%$) \\
\hline
DiffPool \cite{ying2018hierarchical} & 82.28 ($\downarrow 8.69\%$)& 88.02 ($\downarrow 2.32\%$) & 90.11 & 88.83 ($\downarrow 1.42\%$) & 84.50 ($\downarrow 6.23\%$)\\
SortPool \cite{zhang2018end} & 80.70 ($\downarrow 14.24\%$)& 92.33 ($\downarrow 1.88\%$) & 94.10 & 92.14 ($\downarrow 2.08\%$) & 90.35 ($\downarrow 3.99\%$) \\
Top-$k$ Pool \cite{gao2019graph} & 84.31 ($\downarrow 11.81\%$)& 93.53 ($\downarrow 2.17\%$) & 95.60 & 94.44 ($\downarrow 1.21\%$) & 88.02 ($\downarrow 7.93\%$)\\
SAGPool \cite{lee2019self} & 88.08 ($\downarrow 7.82\%$)& 90.86 ($\downarrow 4.91\%$)& 95.55 & 92.22 ($\downarrow 3.49\%$) & 83.99 ($\downarrow 12.10\%$)\\
\hline
GCN  \cite{kipf2016semi} & 84.94 ($\downarrow 6.60\%$)& 89.38 ($\downarrow 1.72\%$) & 90.94 & 90.25 ($\downarrow 0.76\%$) & 86.00 ($\downarrow 5.43\%$) \\ 
\rowcolor{gray!20} GCN + CAL & 89.38 ($\downarrow 6.03\%$) & 93.50 ($\downarrow 1.70\%$) & 95.12  & 95.06 ($\downarrow 0.06\%$)  & 93.31 ($\downarrow 1.90\%$) \\ 
\hline
GIN \cite{xu2018how} & 87.50 ($\downarrow 9.55\%$) & 93.94 ($\downarrow 2.89\%$) & 96.74 & 94.88 ($\downarrow 1.92\%$) & 89.62 ($\downarrow 7.36\%$) \\
\rowcolor{gray!20} GIN + CAL & 93.19 ($\downarrow 3.87\%$) & 96.31 ($\downarrow 0.65\%$) & 96.94 & 96.56 ($\downarrow 0.39\%$) & 95.25 ($\downarrow 1.74\%$)  \\
\hline
GAT \cite{velivckovic2018graph} & 84.62 ($\downarrow 8.71\%$)& 89.50 ($\downarrow 3.44\%$) & 92.69 & 92.31 ($\downarrow 0.41\%$) & 87.62 ($\downarrow 5.47\%$) \\
\rowcolor{gray!20} GAT + CAL & 92.44 ($\downarrow 4.37\%$)& 96.25 ($\downarrow 0.42\%$) & 96.66 & 96.12 ($\downarrow 0.56\%$) & 92.56 ($\downarrow 4.24\%$) \\ 
\bottomrule
\end{tabular}}
\end{table*}

\subsection{Performance on Synthetic Graphs (RQ1)}

\begin{figure}[t]
    \centering
    \includegraphics[width=0.8\linewidth]{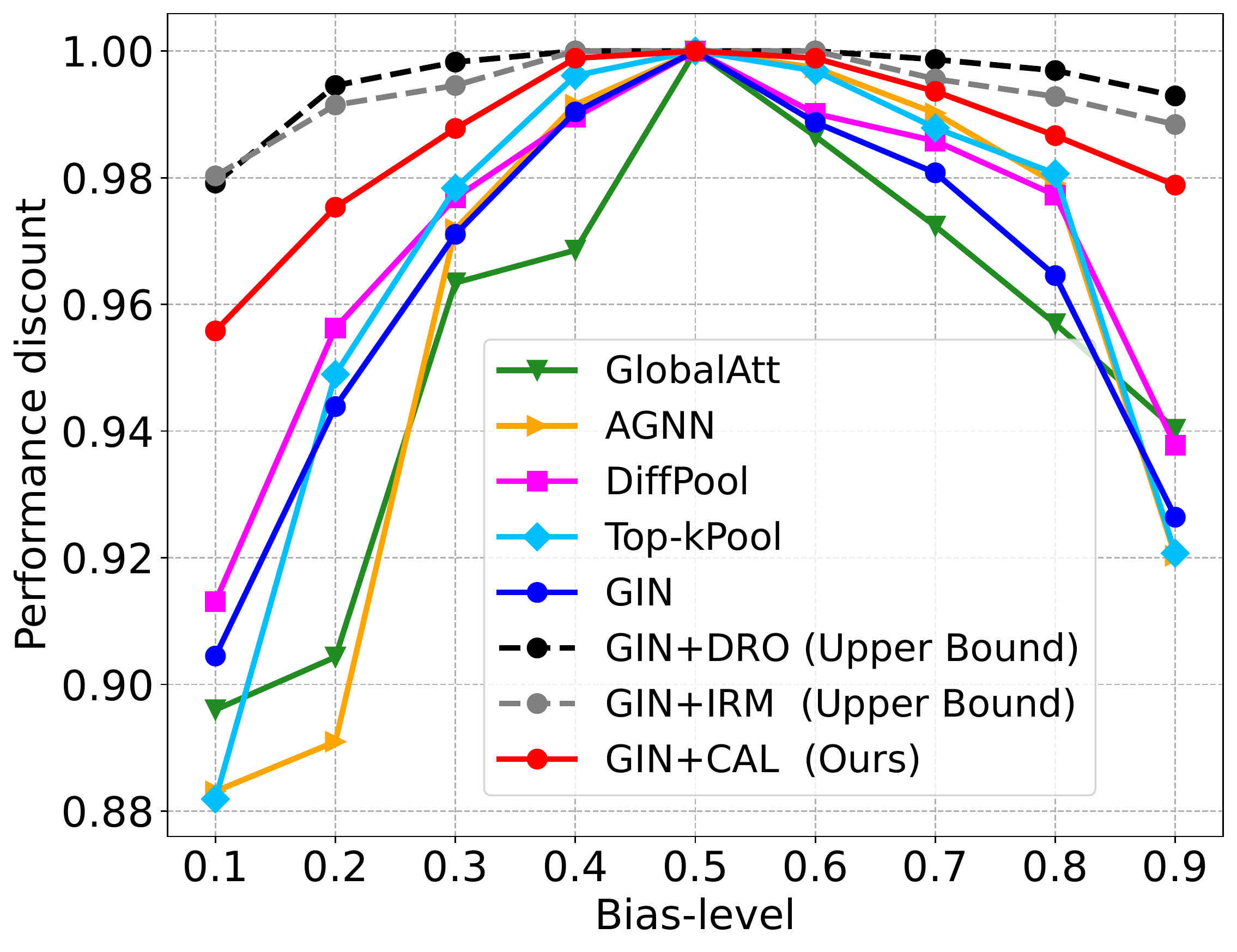}
    \vspace{-4mm}
    \caption{The performance discount on synthetic datasets with different bias-levels.}
    \label{fig:syn_norm_results}
    \vspace{-4mm}
\end{figure}

To explore whether CAL can alleviate the OOD issue, we first conduct experiments on SYN-$b$ with different biases: $b \in \{0.1, 0.2, ..., 0.9\}$.
The experimental results are summarized in Table \ref{table:main_syn} and Figure \ref{fig:syn_norm_results}.
We have the following \textbf{Obs}ervations:

\textbf{Obs 1: Refining discriminative features without considering the causality leads to poor OOD generalization.}
For the unbiased dataset, most attention- and pooling-based baselines, such as GlobalAtt, SuperGAT, SortPool, Top-$k$ Pool, outperform GCN.
It indicates the effectiveness of extracting discriminative features in the ID setting.
However, as the bias-level goes to extremes, the performance dramatically deteriorates.
For instance, the performance drop of attention-based methods ranges from $7.37\%\sim12.75\%$ on SYN-$0.1$, and $3.79\%\sim13.79\%$ on SYN-$0.9$;
Pooling-based methods drop from  $7.82\%\sim14.24\%$ and $3.99\%\sim12.10\%$ for SYN-$0.1$ and SYN-$0.9$.
These indicate that simply extracting discriminative features by attention or pooling module is prone to capture the data biases.
These are also beneficial for reducing the training loss but lead to poor OOD generalization. 
\syd{Taking SYN-$0.9$ as an example, most ``House'' co-occur with ``Tree'' in the training data, so the model will mistakenly learn shortcut features from the ``Tree''-type trivial subgraphs to make predictions, instead of probing the ``House''-type causal subgraphs. 
This will mislead the model to adopt the ``Tree'' pattern to make decisions in the inference stage.}

\textbf{Obs 2: GNNs with better ID performance tend to have worse OOD generalization.}
For the unbiased dataset, GIN achieves the best performance (96.74\%), while GAT (92.69\%) outperforms the GCN (90.94\%).
This indicates that the in-distribution (ID) performance of these models exhibits such an order: GIN $>$ GAT $>$ GCN.
However, when the bias is changed to 0.1 and 0.9, the performance of GIN drops by $9.55\%$ and $7.36\%$, GAT drops by $8.71\%$ and $5.47\%$ and GCN drops by $6.60\%$ and $5.43\%$, respectively.
It shows that the rankings of models' robustness against OOD issues are in the opposite order: GCN $>$ GAT $>$ GIN.
This indicates that GNNs with better ID performance are prone to learn more shortcut features.
Similar trends also occur in other baselines.
After adopting the proposed CAL, this phenomenon is significantly alleviated, which verifies the effectiveness of CAL in overcoming the OOD issue.

\begin{table*}[htb]
\centering
\caption{Test Accuracy (\%) of classification. For TUDataset, we perform 10-fold cross-validation and report the mean and standard derivations. Our methods are highlighted with gray background. If the performance improves, the number is bolded.}
\label{table:main_all}
\setlength{\tabcolsep}{3.1mm}{\begin{tabular}{l|ccc|ccc|cc}
\toprule
Dataset &  MUTAG & NCI1  & PROTEINS & COLLAB & IMDB-B & IMDB-M & MNIST & CIFAR-10  \\ 
\toprule
GK \cite{shervashidze2009efficient}  & 81.58\footnotesize{$\pm$2.11} & 62.49\footnotesize{$\pm$0.27} &  71.67\footnotesize{$\pm$0.55} & 72.84\footnotesize{$\pm$0.28} & 65.87\footnotesize{$\pm$0.98} & 43.89\footnotesize{$\pm$0.38} & - & - \\ 
WL \cite{shervashidze2011weisfeiler} & 82.05\footnotesize{$\pm$0.36} & 82.19\footnotesize{$\pm$0.18} &  74.68\footnotesize{$\pm$0.50} & 79.02\footnotesize{$\pm$1.77} & 73.40\footnotesize{$\pm$4.63} & 49.33\footnotesize{$\pm$4.75} & - & - \\
DGK \cite{yanardag2015deep} 		 & 87.44\footnotesize{$\pm$2.72} & 80.31\footnotesize{$\pm$0.46} & 75.68\footnotesize{$\pm$0.54} & 73.09\footnotesize{$\pm$0.25} & 66.96\footnotesize{$\pm$0.56} & 44.55\footnotesize{$\pm$0.52} & - & - \\
\hline 
GlobalAtt \cite{li2015gated}     & 88.27\footnotesize{$\pm$8.65} & 81.17\footnotesize{$\pm$1.04} &  72.60\footnotesize{$\pm$4.37} & 81.48\footnotesize{$\pm$1.46} & 69.10\footnotesize{$\pm$3.80} & 51.40\footnotesize{$\pm$2.91} & - & - \\
AGNN \cite{thekumparampil2018attention}		 & 79.77\footnotesize{$\pm$8.54} & 79.96\footnotesize{$\pm$2.37} &  75.66\footnotesize{$\pm$3.94} & 81.10\footnotesize{$\pm$2.39} & 73.10\footnotesize{$\pm$4.07} & 49.73\footnotesize{$\pm$3.72} & - & - \\
DiffPool \cite{ying2018hierarchical} & 85.61\footnotesize{$\pm$6.22} & 75.06\footnotesize{$\pm$3.66} &  76.25\footnotesize{$\pm$4.21} & 79.24\footnotesize{$\pm$1.66} & 74.47\footnotesize{$\pm$3.84} & 49.20\footnotesize{$\pm$3.10} & - & -  \\
SortPool \cite{zhang2018end}         & 86.17\footnotesize{$\pm$7.53} & 79.00\footnotesize{$\pm$1.68} & 75.48\footnotesize{$\pm$1.62} & 77.84\footnotesize{$\pm$1.22} & 73.00\footnotesize{$\pm$3.50} & 49.53\footnotesize{$\pm$2.29} & - & - \\
\hline
GCN  \cite{kipf2016semi}             & 88.20\footnotesize{$\pm$7.33} & 82.97\footnotesize{$\pm$2.34} &  75.65\footnotesize{$\pm$3.24} & 81.72\footnotesize{$\pm$1.64} & 73.89\footnotesize{$\pm$5.74} & 51.53\footnotesize{$\pm$3.28} & 90.49 & 54.68 \\ 
\rowcolor{gray!20} GCN + CAL & \textbf{89.24\footnotesize{$\pm$8.72}} & \textbf{83.48\footnotesize{$\pm$1.94}} & \textbf{76.28\footnotesize{$\pm$3.65}} & \textbf{82.08\footnotesize{$\pm$2.40}} & \textbf{74.40\footnotesize{$\pm$4.55}} & \textbf{52.13\footnotesize{$\pm$2.96}} & \textbf{94.58} & \textbf{56.21} \\ 
\hline
GIN \cite{xu2018how} & 89.42\footnotesize{$\pm$7.40} & 82.71\footnotesize{$\pm$1.52} &  76.21\footnotesize{$\pm$3.83} & 82.08\footnotesize{$\pm$1.51} & 73.40\footnotesize{$\pm$3.78} & 51.53\footnotesize{$\pm$2.97} &  96.51 & 56.36 \\
\rowcolor{gray!20} GIN + CAL & \textbf{89.91\footnotesize{$\pm$8.34}} & \textbf{83.89\footnotesize{$\pm$1.93}} &  \textbf{76.92\footnotesize{$\pm$3.31}} & \textbf{82.68\footnotesize{$\pm$1.25}} & \textbf{74.13\footnotesize{$\pm$5.21}} & \textbf{52.60\footnotesize{$\pm$2.36}} &  \textbf{96.93} & \textbf{56.63} \\ 
\hline
GAT \cite{velivckovic2018graph}   & 88.58\footnotesize{$\pm$7.54} & 82.11\footnotesize{$\pm$1.43} &  75.96\footnotesize{$\pm$3.26} & 81.42\footnotesize{$\pm$1.41} & 72.70\footnotesize{$\pm$4.37} & 50.60\footnotesize{$\pm$3.75} &  95.53 & 64.22 \\
\rowcolor{gray!20} GAT + CAL & \textbf{89.94\footnotesize{$\pm$8.78}} & \textbf{83.55\footnotesize{$\pm$1.42}} &  \textbf{76.39\footnotesize{$\pm$3.65}} & \textbf{82.12\footnotesize{$\pm$1.95}} & \textbf{73.30\footnotesize{$\pm$4.16}} & \textbf{50.93\footnotesize{$\pm$3.84}} &  \textbf{95.91} & \textbf{66.16} \\
\bottomrule
\end{tabular}}
\end{table*}

\textbf{Obs 3: Mitigating the confounder achieves more stable performance on OOD datasets.}
We first define the performance discount on SYN-$b$ as the accuracy on SYN-$b$ normalized by the accuracy on unbiased SYN-0.5. 
It indicates the degree of the performance degradation on biased synthetic datasets, without considering the model's ID generalization.
We plot the performance discount curves on SYN-$b$ with $b \in \{0.1, 0.2, ..., 0.9\}$.
As depicted in Figure \ref{fig:syn_norm_results}, we observe that pooling-based methods outperform GIN in a small range of bias-levels ($0.2\sim0.8$), while the performance drops sharply when $b=0.1$ or $0.9$. For example, the performance discount of Top-$k$ Pool drops from 0.95 to 0.88 as $b$ reduces from 0.2 to 0.1.
Attention-based methods perform worse than GIN when $b<0.5$.
For $b>0.5$, AGNN achieves better performance than GIN, while GlobalAttention often performs worse.
These results reflect that attention- or pooling-based methods all have their own weaknesses, such that they cannot consistently overcome the diverse distribution shifts.
Equipped with CAL, GIN (red curve) consistently outperforms all the baselines on all ranges of bias-levels and obviously keeps a large gap, which further demonstrates the significance of mitigating the confounding effect, and the effectiveness of CAL.
For comprehensive comparisons, we also plot two upper bound methods: IRM and DRO (dash lines), which require additional annotation information of trivial subgraphs for training. 
We observe that, even without additional information, CAL achieves comparable performance with these upper bound methods.

\subsection{Performance on Real-world Graphs (RQ2)}
Unlike synthetic graphs, there may not exist visible or specific patterns of the causal/trivial subgraphs in real-world graphs.
However, there still exist irregular core-subgraphs \cite{ReFine, lin2021generative, ying2019gnnexplainer,knyazev2019understanding} that determine the predictions, which will inevitably involve different degrees of biases caused by the complementary parts.
Similar to SYN-$b$, they mislead the GNNs to learn the spurious correlations.
Hence, we verify the practicability of CAL on eight real-world datasets.
We report the results of the baselines from the original papers by default and reproduce the missing results.
The results are summarized in Table \ref{table:main_all} and we make the following \textbf{Obs}ervations:

\textbf{Obs 4: The OOD issue is widespread in real-world datasets.}
Attention-based and pooling-based methods are on a par with GNNs, and they both outperform graph kernel-based methods in most cases.
It can be seen from the last six rows in Table \ref{table:main_all}, when CAL is applied to different GNN models, it consistently produces further performance improvements.
It demonstrates that the distribution shifts also widely exist in real-world datasets.
Specifically, we can find that GCN often performs worse than other GNNs, attention-based or pooling-based methods, while the performance significantly improves after adopting CAL. 
For instance, on IMDB-B and MNIST datasets, GCN+CAL achieves $1.92\%$ and $4.52\%$ relative improvements, respectively. 
This indicates that GCN is vulnerable to the distribution shift in certain datasets. 
Thanks to the causality, CAL will push GCN to pay more attention to causal features, which can establish robustness against the widespread OOD issues and achieve better generalization.

\subsection{Ablation Study (RQ3)}
In this section, we investigate the impact of the node/edge attention, random combination and the loss coefficients $\lambda_1$ and $\lambda_2$.

\begin{figure}[t]
    \centering
    \includegraphics[width=0.8\linewidth]{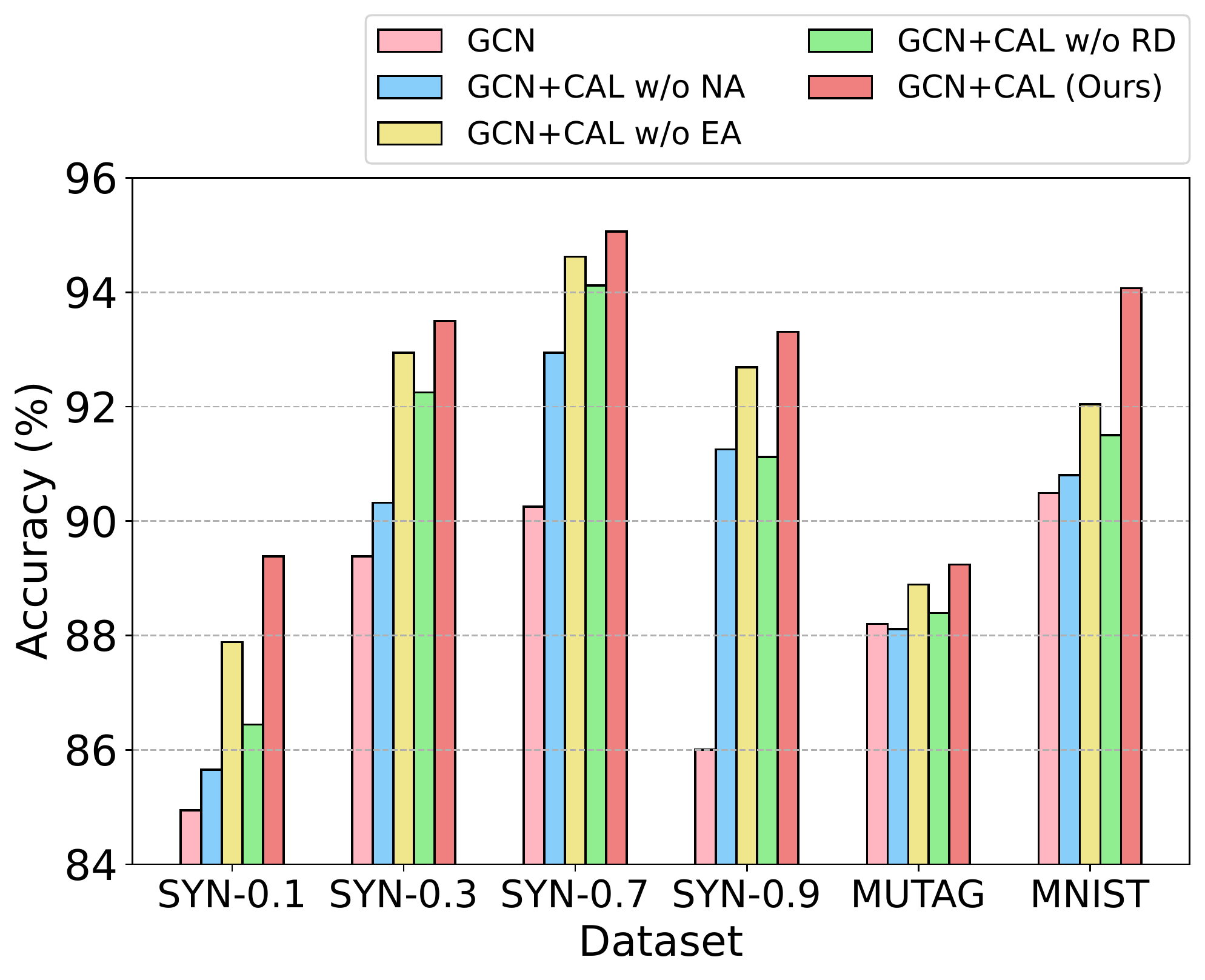}
    \vspace{-4mm}
    \caption{The comparison of different components in CAL.}
    \label{fig:ab1}
    \vspace{-4mm}
\end{figure}

\textbf{Node Attention v.s. Edge Attention.}
Node Attention (NA) and Edge Attention (EA) refine the features from two orthogonal views: node-level and edge-level.
Here we want to examine the effect of adopting NA or EA alone.
We adopt GCN as the encoder to conduct experiments on four biased synthetic datasets and two real-world datasets.
GCN+CAL w/o NA or EA represents the node/edge attention scores in Equation~\eqref{equ:node_att}/\eqref{equ:edge_att} are evenly set as 0.5.
The experimental results are shown in Figure \ref{fig:ab1}. We can find that:
(1) Comparing NA with EA, the performance of CAL without NA is significantly worse than that without EA, which indicates that the node feature contains more significant information compared with graph structure.
(2) Just adopting NA or EA alone still achieves better performance than baselines, which demonstrates that only applying NA or EA can also disentangle the causal/trivial attended-graph and achieve causal intervention to some extent.

\textbf{Random Combination.}
We need to stratify the confounder distribution for causal intervention.
With the random combination, each causal feature will combine with different types of trivial patterns.
To verify its importance, we change the "Random Addition" module in Figure \ref{fig:model} to "Addition", which just adopts the addition operation orderly, and we rename it as ``GCN+CAL w/o RD''.
The experimental results are shown in Figure \ref{fig:ab1}. We can find that:
(1) The performance drops severely compared with GCN+CAL, which demonstrates the importance of the causal intervention.
(2) GCN+CAL w/o RD can also outperform the GCN baselines. 
We conjecture that just implementing disentanglement makes GNN pay more attention to the causal features, which will slightly ignore the data biases or trivial patterns.
These results also reflect that disentanglement and causal intervention will help each other to improve their own effectiveness.

\begin{figure}[t]
    \centering
    \includegraphics[width=1\linewidth]{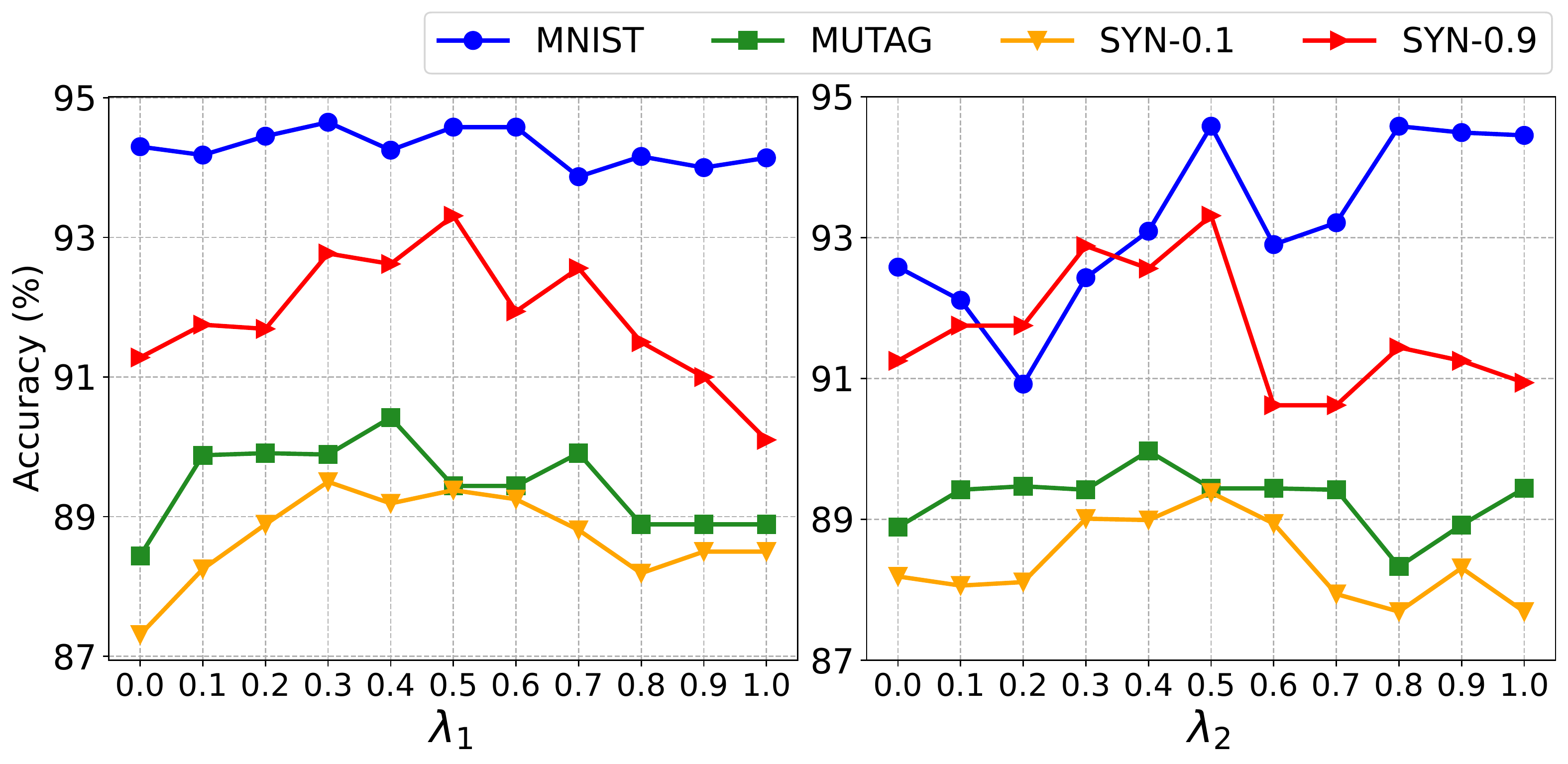}
    \vspace{-8mm}
    \caption{Parameter sensitivity of loss coefficients $\lambda_1$ and $\lambda_2$.}
    \label{fig:ab2}
    \vspace{-8mm}
\end{figure}

\textbf{Loss coefficients $\lambda_1$ and $\lambda_2$.}
According to Equation~\eqref{equ:loss_all}, $\lambda_1$ denotes the strength of the disentanglement for the causal/trivial features, while $\lambda_2$ controls the strength of the causal intervention.
To explore their impacts, we use GCN as the encoder and conduct experiments on two biased synthetic datasets and two real-world datasets.
We fix one coefficient as 0.5 and change the other one in $(0,1)$ with a step size of 0.1.
The experimental results are shown in Figure \ref{fig:ab2}. We can find that:
(1) $\lambda_1$ achieves better performance in a range of $0.3\sim0.7$. Too small or too large values will cause performance degradation.
(2) $\lambda_2$ is not as stable as $\lambda_1$. The optimal range is around $0.3\sim0.5$. 
It leads to a strong decline at $0.5\sim0.8$, which indicates that coefficient $\lambda_2$ should be set prudently.

\subsection{Visualization and Analysis (RQ4)}\label{sec45}

\begin{figure}[t]
    \centering
    \includegraphics[width=0.90\linewidth]{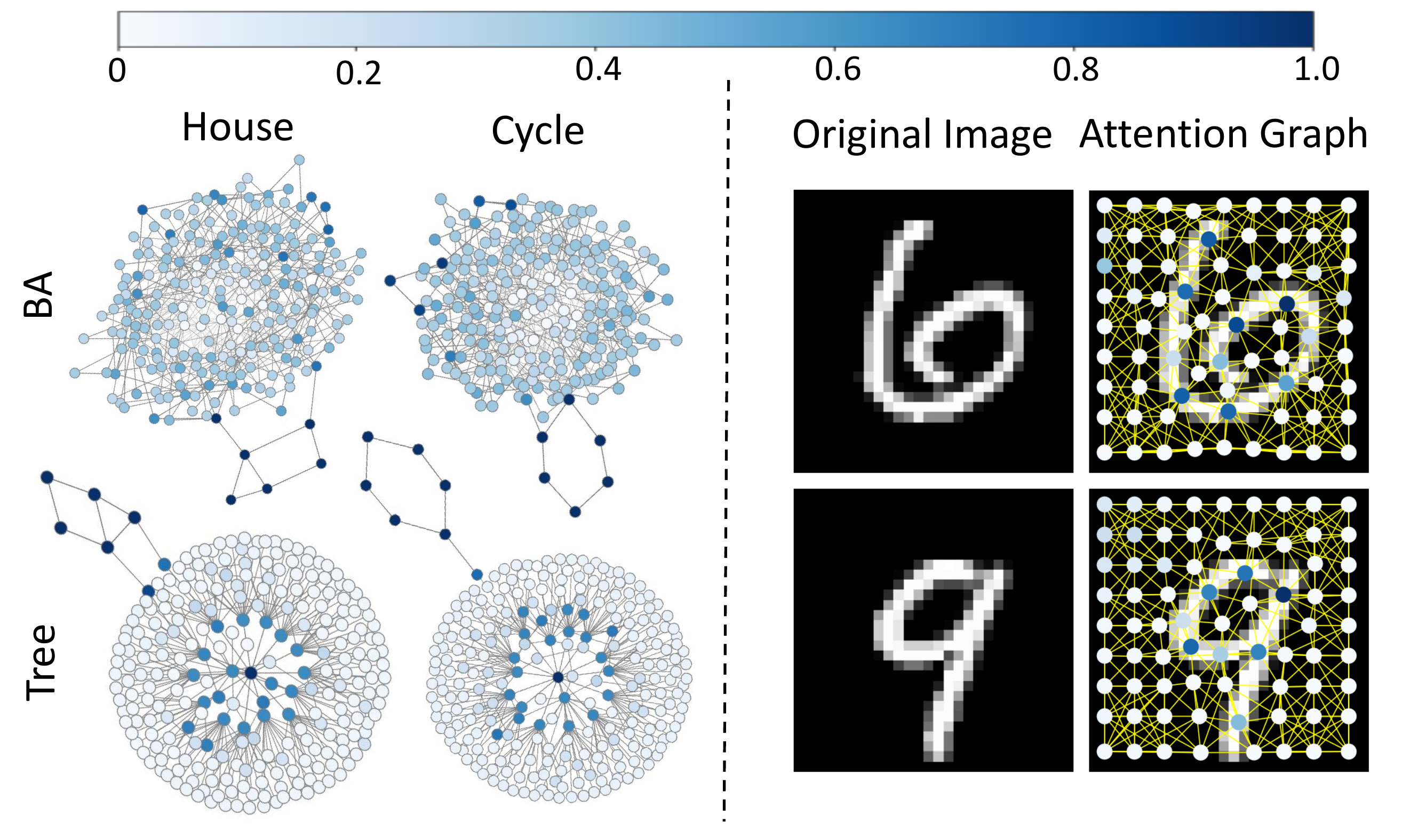}
    \vspace{-4mm}
    \caption{Visualizations of causal attended-graphs. (Left): Synthetic graphs, (Right): MNIST superpixel graphs.}
    \vspace{-4mm}
    \label{fig:syn_vis}
\end{figure}

\textbf{Causal attended-graphs.} 
We plot node/edge attention areas of the causal attended-graphs based on the attention scores in CAL.
We adopt a GCN-based encoder and apply CAL on SYN-$b$ and MNIST superpixel graphs.
The visualizations are shown in Figure \ref{fig:syn_vis}.
Nodes with darker colors and edges with wider lines indicate higher attention scores.
We surprisingly find that almost all the darker colors and wider lines precisely distribute on the deterministic areas, such as the causal subgraphs we defined in the synthetic dataset and the nodes located on digit pixels in MNIST superpixel graphs. 
It further demonstrates that the proposed CAL can effectively capture the causal features with insightful interpretations. 

\begin{figure}[t]
    \centering
    \includegraphics[width=1\linewidth]{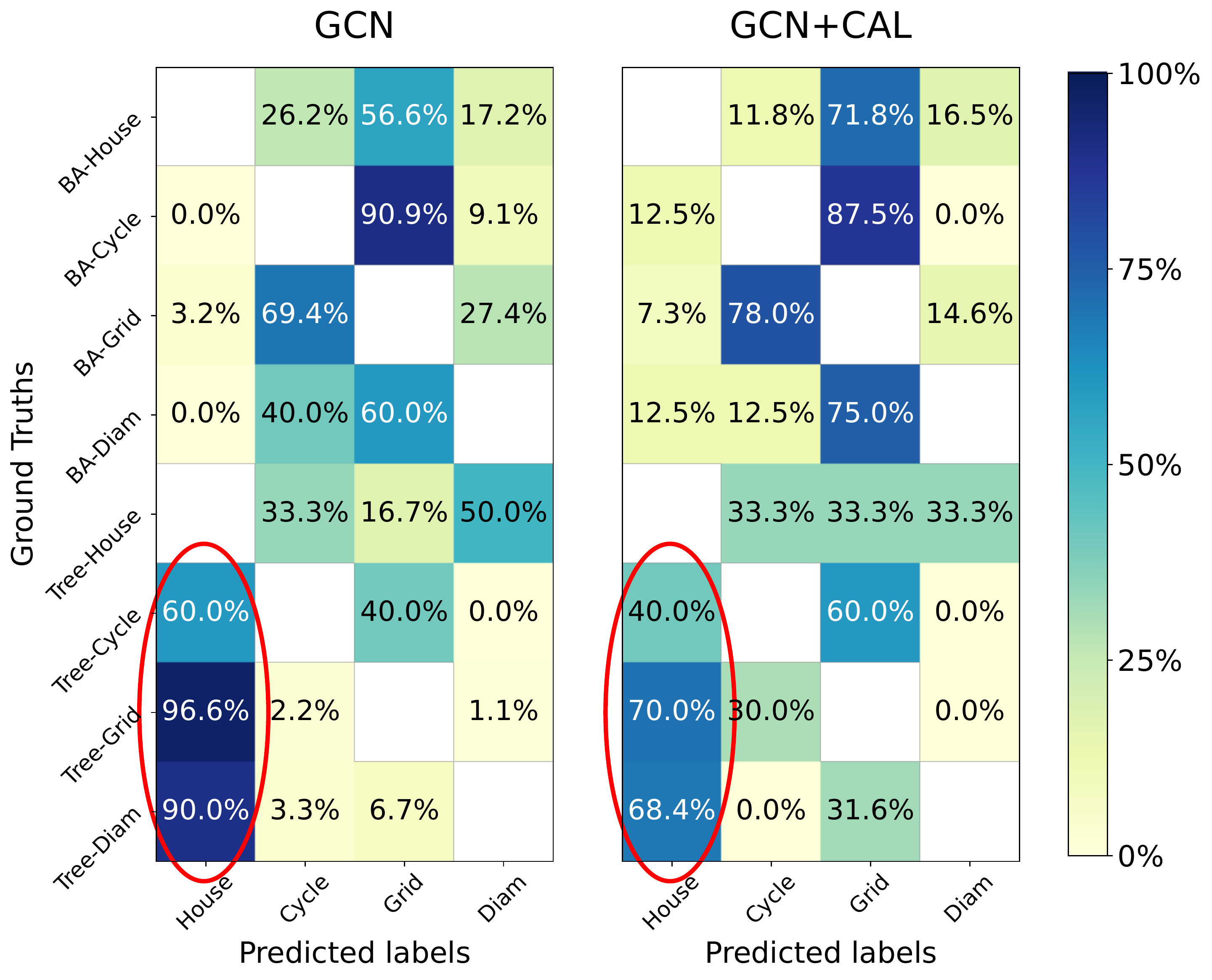}
    \vspace{-8mm}
    \caption{The misclassification distribution.
    Red circle highlights the concentration degree of misclassification.}
    \label{fig:bias_error}
    \vspace{-6mm}
\end{figure}

\noindent\textbf{The explanation for performance improvements.} 
Figure \ref{fig:bias_error} displays the distribution of misclassification on SYN-$b$.
The abscissa represents the predictions, and the ordinate denotes the ground-truth types.
The numbers in each row denote the proportion for each class.
Figure \ref{fig:bias_error} (Left) shows that the wrong predictions of graphs with ``BA'' are mainly distributed in ``Cycle'', ``Grid'' and ``Diamond'' classes, while the wrong predictions of graphs with ``Tree'' mainly concentrate on the ``House'' class (highlighted by the red circle).
On one hand, most of the ``House'' co-occur with ``Tree'' in the training data, GCN tends to capture the shortcut features, \eg ``Tree'' patterns, to make decisions.
Therefore, the other three causal subgraphs with ``Tree'' will mainly be misclassified as ``House'' in the testing set.
On the other hand, only a few ``House'' causal subgraphs co-occur with ``BA'', so the other three causal subgraphs with ``BA'' will almost not be misclassified as ``House''.
In contrast, Figure \ref{fig:bias_error} (Right) shows that, by applying CAL, the concentration of misclassification is obviously alleviated.
This demonstrates that CAL improves performance by mitigating the confounding effect.

\section{Related work}


\textbf{Attention Mechanism} selects the informative features from data, which has obtained great success in computer vision \cite{hu2018squeeze,dosovitskiy2020image,wang2021causal,yang2021causal} and natural language processing tasks \cite{DBLP:conf/nips/VaswaniSPUJGKP17,devlin2019bert}.
In recent years, attention mechanism has gradually become prevalent in the GNN field.
The attention modules for GNNs can be defined over edges \cite{velivckovic2018graph,brody2021attentive,kim2020find,lee2019graph,thekumparampil2018attention} or over nodes \cite{li2015gated,lee2019self,lee2018graph}.
Despite effectiveness, attention learning still stays at how to better fit the statistical correlations between data and labels.
Hence, the learned attentions are inherently biased in OOD settings.
Recent studies \cite{wang2021causal,yang2021causal} propose the causal attention modules to alleviate the bias.
CaaM \cite{wang2021causal} adopts the adversarial training to generate the data partition in each iteration to achieve the causal intervention.
CATT \cite{yang2021causal} proposes in-sample and cross-sample attentions based on front-door adjustment.
However, they are both tailored for computer vision tasks, while cannot transfer to graph learning tasks, due to the irregular and challenging graph-structure data.
Distinct from them, we utilize the disentanglement and causal intervention strategies to strengthen the attention modules for GNNs. 

\noindent\textbf{OOD Generalization} \cite{hendrycks2016baseline,arjovsky2019invariant,rosenfeld2020risks,sagawa2019distributionally} has been extensively explored in recent years.
IRM \cite{arjovsky2019invariant} minimizes the empirical risk under different environments.
Group-DRO \cite{sagawa2019distributionally} adversarially explores the group with the worst risk and achieves generalization by minimizing the empirical risk of the worst group.
Existing efforts \cite{arjovsky2019invariant,rosenfeld2020risks,sagawa2019distributionally} mainly focus on computer vision or natural language processing tasks, while the GNN field is of great need but largely unexplored.
Furthermore, these methods require the environment or group prior information for each training sample, which is expensive in practice.
To alleviate this dilemma, we adopt causal intervention to strengthen the causal relationship between the causal feature and prediction, thereby achieving better generalization.

\noindent\textbf{Causal Inference}s \cite{pearl2000models,pearl2014interpretation} endows the model with the ability to pursue real causality.
A growing number of studies \cite{zhang2020causal,hu2021distilling,tang2020long,niu2021counterfactual} have shown that causal inference is beneficial to diverse computer vision tasks.
CONTA \cite{zhang2020causal} uses backdoor adjustment to eliminate the confounder in weakly supervised semantic segmentation tasks.
DDE \cite{hu2021distilling} proposes to distill the colliding effect between the old and the new data to improve class-incremental learning.
Unlike computer vision, the application of causal intervention in the GNN community is still in its infancy.
CGI \cite{feng2021should} explores how to select trustworthy neighbors for GNN in the inference stage, and demonstrates its effectiveness in node classification.
Recent work \cite{zevcevic2021relating} studies the connection between GNNs and SCM from a theoretical perspective.
\syd{Different from them, we introduce a causal attention learning strategy to mitigate the confounding effect for GNNs.
It encourages GNNs to pay more attention to causal features, which will enhance the robustness against the distribution shift.}

\section{Conclusion}
In this work, we revisit the GNN modeling for graph classification from a causal perspective.
We find that current GNN learning strategies are prone to exploit the shortcut features to support their predictions.
However, the shortcut feature actually plays a confounder role. 
It establishes a backdoor path between the causal feature and the prediction, which misleads the GNNs to learn spurious correlations.
To mitigate the confounding effect, we propose the causal attention learning (CAL) strategy for GNNs.
CAL is guided by the backdoor adjustment from the causal theory.
It encourages the GNNs to exploit causal features while ignoring the shortcut parts.
Extensive experimental results and analyses verify its effectiveness.
Future studies include adopting powerful disentanglement methods and more advanced causal intervention strategies to improve the CAL.
We will also make efforts to apply CAL to other graph learning tasks, such as node classification or link prediction.

\section{Acknowledgments}
This work is supported by the National Key Research and Development Program of China (2020AAA0106000), and the National Natural Science Foundation of China (U19A2079, U21B2026). This research is also supported by CCCD Key Lab of Ministry of Culture and Tourism and Sea-NExT Joint Lab.

\bibliographystyle{ACM-Reference-Format}
\bibliography{main}

\appendix

\clearpage

\section{Appendix}

\subsection{Algorithm}\label{app1}
We provide the detailed implementation of the proposed casual attention learning (CAL) in Algorithm \ref{alg1}.
We adopt the causal attended-graph for prediction in the inference stage.

\begin{algorithm}[htb]
\caption{Casual Attention Learning} 
\label{alg1}
\begin{algorithmic}[1]
\SetKwInOut{Input}{Input}
\SetKwInOut{Output}{Output} 
\REQUIRE{Dataset $\mathcal{D}$, $f(\cdot)$, attention modules, classifiers, $\lambda_1$, $\lambda_2$} 
\ENSURE{The trained parameters.}
\FOR {sampled $M$ graphs $\{\mathcal{G}^k=\{\Mat{A}^k, \Mat{X}^k\}\}_{k=1}^M$}
\FOR {$k\leftarrow 1$ \textbf{to} $M$}
\STATE $\Mat{H}^k \leftarrow f(\Mat{A}^k, \Mat{X}^k\})$
\STATE Compute $\alpha_{c_i}, \alpha_{t_i}\leftarrow$ Equation~\eqref{equ:node_att} for all nodes of $\mathcal{G}^k$
\STATE Compute $\beta_{c_{ij}}, \beta_{t_{ij}}\leftarrow $ Equation~\eqref{equ:edge_att} for all edges of $\mathcal{G}^k$
\STATE Get masks $\Mat{M}_a^k$ and $\Mat{M}_x^k$ based on $\beta_{c_{ij}}$ and $\alpha_{c_i}$
\STATE Get masks $\overline{\Mat{M}_a^k}$ and $\overline{\Mat{M}_x^k}$ based on $\beta_{t_{ij}}$ and $\alpha_{t_i}$
\STATE $\mathcal{G}_c^k\leftarrow \{\Mat{A}^k\odot \Mat{M}_a^k, \Mat{X}^k\odot \Mat{M}_x^k\}$ \tcp{causal}
\STATE $\mathcal{G}_t^k\leftarrow \{\Mat{A}^k\odot \overline{\Mat{M}_a^k}, \Mat{X}^k\odot \overline{\Mat{M}_x^k}\}$ \tcp{trivial}
\STATE $\Mat{h}_{\mathcal{G}_{c}^k}, \Mat{z}_{\mathcal{G}_c^k} \leftarrow$ Equation~\eqref{equ:zc}
\STATE $\Mat{h}_{\mathcal{G}_{t}^k}, \Mat{z}_{\mathcal{G}_t^k} \leftarrow$ Equation~\eqref{equ:zt}
\ENDFOR
\STATE Supervised loss: $\mathcal{L}_{\mathrm{sup}} \leftarrow$ Equation~\eqref{equ:loss1}
\STATE Uniform loss: $\mathcal{L}_{\mathrm{unif}} \leftarrow$ Equation~\eqref{equ:loss2}
\STATE $I\leftarrow\ \mathrm{Shuffle}([1,...,M])$ 
\FOR {$k\leftarrow 1$ \textbf{to} $M$}
\STATE $i\leftarrow I[k]$
\STATE  $\Mat{h}_{\mathcal{G}^k} \leftarrow \Mat{h}_{\mathcal{G}_{c}^k} + \Mat{h}_{\mathcal{G}_{t}^i}$ \tcp{random combination}
\STATE  $\Mat{z}_{\mathcal{G}^k} \leftarrow \mathrm{\Phi}(\Mat{h}_{\mathcal{G}^{k}})$
\ENDFOR
\STATE Causal loss: $\mathcal{L}_{\mathrm{caus}} \leftarrow$ Equation~\eqref{equ:loss3}
\STATE Total loss: $\mathcal{L} \leftarrow \mathcal{L}_{\mathrm{sup}} + \lambda_1\mathcal{L}_{\mathrm{unif}} + \lambda_2\mathcal{L}_{\mathrm{caus}}$
\STATE Update all the trainable parameters to minimize $\mathcal{L}$
\ENDFOR
\end{algorithmic}
\end{algorithm}

\subsection{Datasets Details}\label{app2}
In this section, we give more details about the synthetic datasets and real-world datasets.

\textbf{1) Synthetic graphs.} 
For each synthetic graph instance, it consists of two subgraphs: trivial and critical subgraphs. 
We introduce the proposed trivial subgraph and critical subgraph as follows:
\begin{itemize}[leftmargin=*]
\item \textbf{Trivial subgraph}. There exist two types of trivial subgraphs: BA-SHAPES and Tree.
The BA-SHAPES is a Barabási-Albert (BA) graph \cite{barabasi1999emergence}, and we abbreviate it as ``BA'' in this paper.
The ``Tree'' graph is a base 12-level balanced binary tree \cite{ying2019gnnexplainer}.
To reduce the influence, we control the number of nodes in the two kinds of trivial subgraphs to be similar.
\item \textbf{Causal subgraph}. There are four types of causal subgraphs: ``House'', ``Cycle'', ``Grid'', ``Diamond''. The visualizations of these trivial subgraphs and causal subgraphs are depicted in Figure \ref{fig:syn_dataset}.
\end{itemize}
For each synthetic graph instance, a causal subgraph is randomly attached on one node of a trivial subgraph. 
Then the resulting graph is further perturbed by adding 10\% random edges.
We take the one-hot form of the node degree as the node feature and set the dimension of node feature to 20.
The synthetic graph examples are displayed in Figure \ref{fig:syn_dataset}. The statistics of the synthetic datasets are summarized in Table \ref{table:dataset_s}. 
We split the dataset into training, validation and testing set with the ratio of 7: 1: 2.

\begin{table}[htb]
\centering
\vspace{-2mm}
\caption{Statistics of datasets used in experiments.}
\vspace{-2mm}
\label{table:dataset_s}
\setlength{\tabcolsep}{2.5mm}{\begin{tabular}{l|cccc}
\toprule
\textbf{Dataset} & \#\textbf{Graphs} & \#\textbf{Nodes} & \#\textbf{Edges} & \#\textbf{Classes} \\ 
\toprule
SYN-$b$ & 8000 & 230$\sim$247 & 542$\sim$1000 & 4 \\
\hline
MUTAG & 188 & 17.93 & 19.79 & 2 \\
NCI1 & 4110 & 29.87 & 32.30 & 2 \\
PROTEINS & 1113 & 39.06 & 72.82 & 2 \\
\hline
COLLAB & 5000 & 74.49 & 2457.78 & 3\\
IMDB-B & 1000 & 19.77 & 96.53 & 2 \\
IMDB-M & 1500 & 13.00 & 65.94 & 3\\
\hline
MNIST & 70000 & 70.57 & 564.66 & 10 \\
CIFAR-10 & 60000 & 117.63 & 941.04 & 10 \\
\bottomrule
\end{tabular}}
\end{table}

\textbf{2) Real-world graphs.} 
To demonstrate the practicality of the proposed CAL, we conduct experiments on TUDataset \cite{morris2020tudataset} and Superpixel graphs \cite{knyazev2019understanding}.
For TUDataset, we gather three biological datasets (MUTAG, NCI1, PROTEINS) and three social networks datasets (COLLAB, IMDB-B, IMDB-M), which are commonly used in graph classification benchmarks \cite{xu2018how, dwivedi2020benchmarking}.
Following \cite{dwivedi2020benchmarking,xu2018how,ying2018hierarchical}, we use 10-fold cross-validation and report average accuracy and standard deviation.
The superpixel graphs \cite{knyazev2019understanding, dwivedi2020benchmarking} includes MNIST and CIFAR-10, which are classical image classification datasets converted into graphs using superpixels technology \cite{achanta2012slic} and assigning each node's features as the superpixel coordinates and intensity.
Following \cite{dwivedi2020benchmarking, knyazev2019understanding}, we split the MNIST and CIFAR-10 to 55K training/5K validation/10K testing, and 45K training/5K validation/10K testing, respectively.
All the detailed statistics about the real-world datasets are summarized in Table \ref{table:dataset_s}.

\subsection{Hyper-parameters}\label{app3}
As for training parameters, we train the models for 100 epochs with batch size of 128. 
We optimize all models with the Adam optimizer.
For SYN-$b$ and TUDataset, we use GCN, GIN and GAT as GNN encoders with 3 layers and 128 hidden units.
For Superpixel graphs MNIST and CIFAR-10, we use the GNN encoders with 4 layers and 146 hidden units as \cite{dwivedi2020benchmarking}.
For all the baselines, we follow the default settings from original papers and reproduce the missing results.
For the proposed CAL, we search $\lambda_1$ and $\lambda_2$ in $(0.1, 1.0)$ with a step size of $0.1$ and report the results with the best settings.
We adopt NVIDIA 2080 Ti (11GB GPU) to conduct all our experiments, the training time comparison is shown as Table \ref{table:time}.

\begin{table}[htb]
\centering
\vspace{-2mm}
\caption{Training time (minutes) comparison.}
\vspace{-2mm}
\label{table:time}
\setlength{\tabcolsep}{1.8mm}{\begin{tabular}{l|ccccc}
\toprule
Method & SYN-$b$ & MUTAG & NCI1 & IMDB-M & MNIST \\ 
\toprule
GCN         & 4.16 & 1.03 & 12.71 & 4.61 & 57.20  \\
GCN + CAL   & 6.67 & 1.35 & 17.37 & 6.16 & 75.80 \\
\bottomrule
\end{tabular}}
\end{table}

\end{document}